\documentclass{article} 
\usepackage[final]{colm2026_conference} 

\usepackage{microtype}
\usepackage{hyperref}
\usepackage{url}
\usepackage{booktabs}

\usepackage{lineno}

\usepackage{comment}
\usepackage{multirow}
\usepackage{graphicx}
\usepackage{subcaption}
\usepackage{enumitem}
\usepackage{makecell}
\usepackage{wrapfig}
\usepackage{threeparttable}

\definecolor{darkblue}{rgb}{0, 0, 0.5}
\hypersetup{colorlinks=true, citecolor=darkblue, linkcolor=darkblue, urlcolor=darkblue}

\title{MoLGE: Mixture of Language Group Experts for Efficient Scaling of Massively Multilingual Speech Recognition}

\author{Sangmin Lee, Woojin Chung, Woongjib Choi, and Hong-Goo Kang \\
Dept. of Electronics and Electrical Engineering, Yonsei University, Seoul, South Korea\\
\texttt{\{sangmin\_lee,woojinchung,woongzip1\}@dsp.yonsei.ac.kr, hgkang@yonsei.ac.kr}}

\begin{document}

\ifcolmsubmission
\linenumbers
\fi

\maketitle

\begin{abstract}

Massively multilingual automatic speech recognition (ASR) models covering hundreds of languages must maintain robust performance across diverse linguistic and acoustic conditions. 
However, these models often encounter the \textbf{\textit{curse of multilinguality}}, where model capacity is diluted across languages.
To address this challenge, we propose Mixture of Language Group Experts (MoLGE), built upon speech self-supervised models (S3Ms). MoLGE assigns dedicated expert modules to clusters of similar languages, reducing the number of required submodules compared to conventional language-specific Mixture-of-Experts (MoE) schemes. 
It further integrates a hierarchical Low-Rank Adaptation (LoRA) strategy into the disentangled acoustic and linguistic components of the S3M architecture, enabling efficient modeling of language-specific characteristics while maintaining parameter efficiency.
Further, we investigate the impact of language grouping strategies based on both linguistic and data-driven criteria on overall performance, providing an interpretable perspective on how language structure influences scalability in multilingual speech systems.
In experiments, we evaluate MoLGE on a multilingual benchmark encompassing 495 languages. Results demonstrate that MoLGE consistently outperforms dense multilingual baselines with a minimal increase in trainable parameters. 
Notably, these language grouping strategies yield substantial improvements for both phonetic and orthographic aspects of ASR modeling.
Our findings suggest that structured language specialization provides an effective pathway for massively scaling language coverage of multilingual ASR.
    
\end{abstract}
\section{Introduction}

Scaling automatic speech recognition (ASR) to hundreds of languages within a unified model remains a core challenge in multilingual speech processing. While these systems aim for broad accessibility, they must navigate significant linguistic and acoustic diversity. Consequently, performance often degrades as the number of supported languages increases, a phenomenon primarily driven by limited model capacity relative to the task's complexity.

Since the advent of speech self-supervised models (S3Ms)~\citep{wav2vec2,hubert,wavlm}, speech representation learning has advanced significantly by leveraging large-scale unlabeled corpora. These models have achieved robust performance across various downstream tasks, including ASR, in some cases surpassing human-level transcription accuracy on monolingual benchmarks.
Consequently, recent research has increasingly focused on extending S3Ms to multilingual settings~\citep{xlsr53, xlsr, mhubert, mms, xeus}. Current approaches primarily rely on scaling training data to millions of hours~\citep{usm} and increasing model capacity to billions of parameters~\citep{omniasr} to improve multilingual coverage and cross-lingual generalization.

Despite these advances, significant challenges remain. 
Increasing dataset size and model capacity is computationally demanding and often impractical at scale~\citep{lincom,k2ssl}. 
Recent massively multilingual systems already require hundreds of millions to billions of parameters, rendering both pretraining and adaptation resource-intensive, particularly in computationally constrained settings~\citep{suslearn}.

Furthermore, multilingual speech models are typically trained on highly imbalanced datasets dominated by high-resource languages.
This imbalance exacerbates the curse of multilinguality~\citep{com1,com2}, as fixed model capacity is shared across many languages while being disproportionately optimized for a small subset. 
Consequently, a single dense multilingual model often underperforms relative to specialized monolingual systems~\citep{interf1,interf2}, highlighting the limitations of uniform scaling for multilingual speech modeling.
These limitations reveal a fundamental tension between language scalability and resource accessibility. While scaling multilingual models aims to support underrepresented languages, their increasing computational demands driven by model size and training data limit their practical utility in the settings they are intended to serve~\citep{opportunity,llmforevery,languagegap}.

To this end, we argue that this tension can be alleviated by shifting from uniform capacity scaling to structured architectural design informed by language relationships. 
Languages are not independent entities; they share phonological, morphological, and genealogical properties that define a structured space of linguistic similarity~\citep{language,uriel}.
Exploiting this structure as an architectural prior allows model capacity to be allocated in a linguistically principled manner, providing a more efficient pathway for multilingual scaling than exhaustive parameter or data expansion.

Building on this motivation, we propose \textit{\textbf{Mixture of Language Group Experts (MoLGE)}}, a parameter-efficient and language-aware framework for massively multilingual speech modeling.
MoLGE enables structured language specialization by incorporating linguistic relationships directly into its expert allocation mechanism, without relying solely on scaling.

Specifically, our framework is built on two design principles.
First, motivated by evidence that lower S3M layers capture universal acoustic features while higher layers encode language-specific information~\citep{layerwise1,layerwise2}, we adopt a speech-oriented adaptation strategy. This includes hierarchical adaptation, where shared LoRA modules model universal acoustics in lower layers and language group-specific experts capture linguistic characteristics in higher layers, as well as shared experts and an attentive pooling router to improve acoustic modeling and stabilize expert specialization. 

Second, we explore language grouping as a mechanism for injecting linguistic prior knowledge into the model architecture. 
We hypothesize that linguistically or acoustically similar languages share structural patterns, and that grouping them under shared experts effectively increases the accessible training signal for each expert, acting as a form of implicit data augmentation.
To this end, we investigate six grouping strategies spanning implicit, embedding-guided, and knowledge-guided approaches, providing a comprehensive evaluation of how different types of linguistic prior influence expert specialization.

Beyond the architectural design, we conduct a systematic analysis of when and why language-aware modeling is most beneficial. 
By evaluating MoLGE across two ASR models with distinct output characteristics, phonetic (language-agnostic Romanized transcription) and orthographic (language-specific scripts), we examine how the complexity of the target linguistic space modulates the effectiveness of structured language grouping.

Our contributions are summarized as follows:
\begin{itemize}[leftmargin=*]
    \item We propose MoLGE, an efficient and scalable framework for massively multilingual ASR that incorporates language structure as an architectural prior for expert specialization.
    \item We integrate speech modeling principles into LoRA and MoLE, incorporating hierarchical adaptation, shared experts, and an attentive pooling router to improve acoustic modeling and expert specialization.
    \item We provide a systematic study of six language grouping strategies across 495 languages, showing that explicit linguistic priors consistently outperform implicit or random priors, and that their benefit scales with the complexity of the target linguistic space.
\end{itemize}

\section{Related Work}
\subsection{Multilingual Speech Foundation Models}
Multilingual speech foundation models have primarily advanced by scaling model capacity and training data to support an increasing number of languages. Early research, such as XLSR-53~\citep{xlsr53}, demonstrated that multilingual self-supervised pretraining with wav2vec 2.0~\citep{wav2vec2} improves cross-lingual transfer and generalization for ASR.
Subsequent models, including XLS-R~\citep{xlsr}, MMS~\citep{mms}, mHuBERT-147~\citep{mhubert}, AfriHuBERT~\citep{afrihubert}, XEUS~\citep{xeus}, and Omnilingual ASR~\citep{omniasr}, have extended this paradigm by scaling to hundreds or thousands of languages and increasing model capacity to the billion-parameter regime.
In parallel, large-scale supervised Whisper-style approaches~\citep{whisper,owsm} have achieved robust multilingual ASR performance by training on massive labeled datasets.
Despite these advances, existing methods predominantly rely on uniform scaling of model size and training data, with limited attention to language-aware modeling that explicitly accounts for linguistic diversity.

\subsection{Scalable Parameter-Efficient Finetuning}
Parameter-efficient fine-tuning (PEFT) methods have been widely adopted to adapt large pretrained models without updating the entire parameter set. Approaches such as adapters~\citep{sequentialadapter,adapterfusion} and Low-Rank Adaptation (LoRA)~\citep{lora} enable efficient task-specific adaptation by introducing a minimal number of trainable parameters. These methods have been successfully applied across both natural language processing (NLP) and speech domains to maintain performance while reducing computational overhead.
In multilingual speech modeling, PEFT has been used to scale models across numerous languages. For example, \cite{mms} employs language-specific adapters to extend dense multilingual models to thousands of languages. However, assigning individual adapters to every language becomes impractical as the scale increases.

To improve scalability, Mixture-of-Experts (MoE)~\citep{moe} approaches introduce conditional computation by activating only a subset of parameters for each input. Recent research further integrates MoE with PEFT, such as Mixture of LoRA Experts (MoLE)~\citep{mole}, enabling expert specialization through lightweight adaptation modules. Nevertheless, existing MoE-based multilingual methods typically allocate experts at the granularity of individual languages~\citep{moie,molange,hdmole}, which limits their scalability in massively multilingual scenarios.

Recent studies in NLP have begun exploring language grouping within MoE frameworks~\citep{medmolge,gthens,thormoe,mol}, suggesting that structured parameter sharing across related languages improves scalability. However, such approaches remain underexplored in massively multilingual speech modeling. Existing methods typically apply language-specific LoRA modules to only a small subset of languages~\citep{lrmoe,maslora,langmole} or focus on specialized scenarios such as code-switching~\citep{laemoe,dlgmoe}, limiting their applicability to large and diverse language sets.

\subsection{Position of Our Work}
Our work departs from prior approaches in two key dimensions. 

First, instead of relying on uniform scaling, we integrate PEFT with speech-aware adaptation and language-group-based routing to enable structured, scalable specialization for a massive set of languages. 
Second, while prior group-aware routing has primarily focused on textual inputs, we address the unique challenges of language grouping from a speech-oriented perspective by conducting an in-depth analysis of various language grouping strategies.

Taken together, we present an efficient scaling method for massively multilingual speech recognition, along with the first systematic study of language grouping and its impact.
\begin{figure*}[!t]
    \centering
    \includegraphics[width=0.98\linewidth]{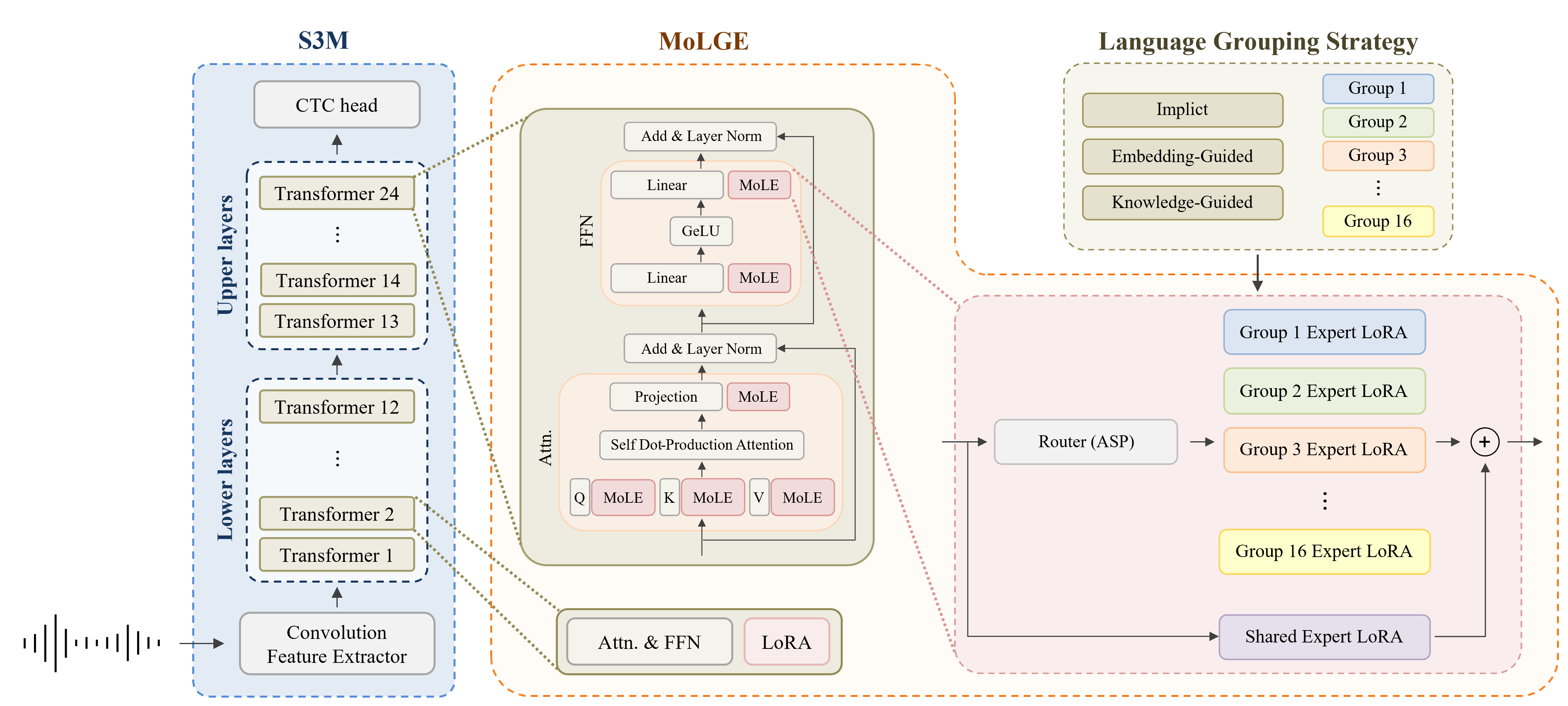}
    \caption{Overall architecture of MoLGE. The conventional S3M is decomposed into lower acoustic and upper linguistic layers, each with distinct adaptation strategies.}
    \label{fig:overall_fig}
    \vspace{-10pt}
\end{figure*}
\section{Proposed Method}
We investigate the role of language grouping in multilingual speech recognition through two primary hypotheses. 
First, grouping linguistically or acoustically similar languages enables experts to capture common structural characteristics, improving parameter efficiency and representation quality, particularly in massively multilingual settings.
Second, under a fixed model capacity (i.e., a constrained number of experts), explicit language grouping strategies based on linguistic or statistical priors facilitate more effective expert specialization than implicit, router-driven grouping learned solely from data.

\subsection{Mixture of Language Group Experts}
In this section, we present \textit{\textbf{Mixture of Language Group Experts (MoLGE)}}, a speech-centric MoLE framework designed for efficient expert specialization through structured language grouping. In contrast to conventional approaches that allocate experts at the individual language level, MoLGE clusters related languages and assigns shared experts to each group. This strategy significantly improves scalability in massively multilingual settings. The overall architecture of MoLGE is illustrated in Figure~\ref{fig:overall_fig}.

We adopt a hierarchical modeling approach based on the layer-wise functional specialization of speech encoders, where lower layers primarily encode language-agnostic acoustic features and higher layers capture language-dependent linguistic representations~\citep{layerwise1, layerwise2}.
Following this observation, we apply \textit{shared LoRA} modules to the lower layers to model global acoustic variability. Since acoustic characteristics across human speech are largely transferable, a unified parameter space efficiently captures cross-lingual acoustic factors without introducing routing overhead. 
Conversely, we deploy \textit{Mixture of LoRA Experts (MoLE)} exclusively in the upper layers. This design allocates expert capacity to the linguistic representation space, where language-specific disambiguation is most critical, enabling efficient modeling of diverse multilingual environments.

To further adapt the MoLE framework for speech recognition, we introduce two key modifications.
Unlike text, speech signals encode both linguistic content and paralinguistic information, such as speaking rate and speaker identity.
We introduce a \textit{shared expert} to capture these paralinguistic factors, which remain largely invariant across different languages. This allows the language-specific experts to focus more effectively on distinct linguistic representations.
In addition, we employ an \textit{Attentive Statistical Pooling (ASP) classifier}~\citep{asp} as the routing mechanism. Given that speech signals often contain temporally imbalanced information, such as long silences, ASP utilizes attention-weighted temporal pooling to emphasize salient acoustic regions while suppressing irrelevant frames. This ensures the router identifies more informative representations for expert assignment.

\begin{figure*}[!t]
    \centering
    \includegraphics[width=0.98\linewidth]{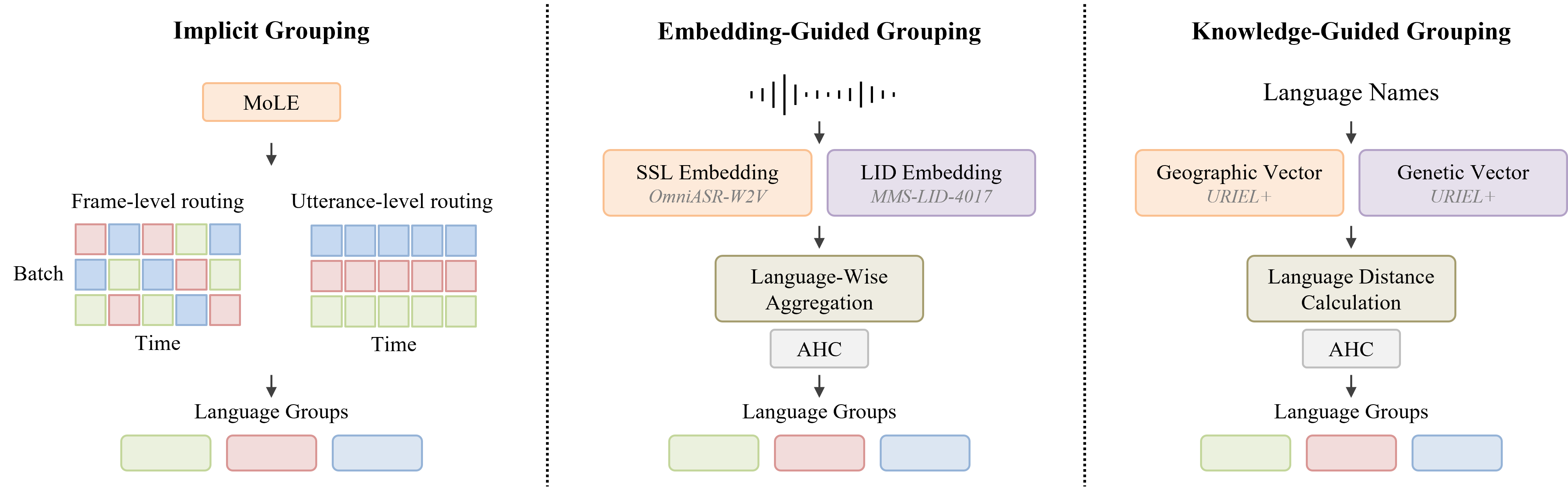}
    \caption{Illustration of the overall procedure for language grouping, covering implicit, embedding-guided, and knowledge-guided strategies.}
    \label{fig:grouping}
    \vspace{-10pt}
\end{figure*}
\subsection{Language Grouping Strategy}
\label{sec:langgroup}
In this section, we construct six language grouping strategies based on diverse information sources. We consider three categories: \textit{implicit}, \textit{embedding-guided}, and \textit{knowledge-guided}, each comprising two variants. The detailed grouping procedure is illustrated in Figure~\ref{fig:grouping}.

Following the principles of statistical cluster analysis~\citep{examination}, the optimal number of clusters can be approximated as $k \approx \sqrt{N/2}$, where $N$ denotes the number of data points. Given our experimental setting of approximately 500 languages ($N \approx 500$), this heuristic suggests $k \approx 15.8$. Accordingly, we fix the number of experts to 16 across all strategies. This choice provides a practical trade-off between model capacity and computational efficiency in massively multilingual scenarios.

\noindent\textbf{Implicit Grouping.}
Implicit grouping refers to a strategy in which languages are not assigned to experts via predefined labels; instead, groupings emerge naturally from the learned behavior of the router. This represents the foundational form of MoE and has been widely studied in the literature~\citep{gshard,switchtf,mixtral}.

Under this strategy, we evaluate two distinct routing granularities: frame-level and utterance-level routing~\citep{moedesign}. Frame-level routing treats each timestep of the speech embedding as an individual token, operating analogously to token-level routing in MoE models in NLP. It allows for fine-grained, dynamic expert assignment throughout the duration of a signal. In contrast, utterance-level routing treats an entire utterance as a single unit, routing the complete sequence to a specific expert via a shared router.
This ensures consistency in expert application across the temporal dimension of the speech input.

\noindent\textbf{Embedding-Guided Grouping.}
Embedding-guided grouping derives language clusters from high-dimensional representations generated by pretrained speech foundation models. Under this strategy, we first extract speech embeddings and construct language prototype vectors by aggregating these representations at the language level. Then, we apply Agglomerative Hierarchical Clustering (AHC) with Ward’s linkage~\citep{ahc} to partition the languages. This clustering approach is motivated by the inherently hierarchical and phylogenetic structure of human languages~\citep{language,langtree1,langtree2,langtree3}. Moreover, Ward’s method minimizes within-cluster variance to encourage the formation of balanced and compact clusters. This structural property is critical for ensuring stable load balancing and mitigating collapse during training.

To evaluate the impact of different representational biases, we implement this strategy using two distinct pretrained speech models. The first variant utilizes OmniASR-W2V~\footnote{https://huggingface.co/facebook/omniASR-W2V-1B}, a multilingual wav2vec2.0-style model trained on 1,239 languages. Because this model is optimized with a general self-supervised learning (SSL) objective, it captures diverse speech characteristics, including speaker, phonetic, and language-level information. 
The second variant employs MMS-LID-4017~\footnote{https://huggingface.co/facebook/mms-lid-4017}, a language identification model which trained on and is capable of classifying 4,017 languages. In contrast to the pure SSL model, MMS-LID-4017 is specifically optimized for language identification (LID), thereby emphasizing features that distinguish one language from another. By comparing these two architectures, we evaluate two routing paradigms: SSL-embedding- and LID–embedding–based routing, respectively.

\noindent\textbf{Knowledge-Guided Grouping.}
Knowledge-guided grouping constructs language clusters by leveraging established linguistic priors. To provide a principled and quantitative representation of this knowledge, we utilize \textit{URIEL+} language vectors~\citep{uriel,urielplus}. These vectors integrate curated linguistic databases into language-level representations, covering 7000+ languages. It also incorporates associated distance metrics derived from a multidimensional attribute space encompassing geographical proximity, typological features, morphological structures, and script-related properties.

Specifically, we construct language distance matrices using two distinct categories of URIEL+ vectors: \textit{geographic} and \textit{genetic}. Geographic vectors capture spatial proximity between languages, capturing regional contact and shared acoustic environments. In contrast, genetic vectors encode genealogical relationships through language family and subfamily hierarchies. This design facilitates typology-aware grouping grounded in shared linguistic ancestry and historical evolution rather than surface-level feature similarity. Consistent with the embedding-guided approach, final language clusters are obtained by applying AHC with Ward's linkage to the resulting distance matrices.

\subsection{Two-Stage Dense-to-Sparse Upcycling}
We apply a two-stage upcycling strategy to the S3M-based ASR models, inspired by LLM upcycling approaches that transform dense models into sparse MoE architectures~\citep{upcycling1,upcycling2,upcycling3}.
This approach allows us to preserve the rich representations learned by dense models while enabling efficient scaling, reducing training cost, and improving parameter utilization in sparse settings.
In the first stage, a dense ASR model is trained on top of pretrained S3M to establish robust global representations and transcription capabilities across languages.
In the second stage, we introduce the MoLGE framework to enable expert specialization across predefined language groups, upcycling the dense model into a sparse, parameter-efficient architecture while further improving performance.
During the second stage, we apply LoRA to all linear layers, including attention projections and feed-forward networks (FFNs). Motivated by prior work suggesting that FFNs encode model knowledge~\citep{ffn1,ffn2,ffn3} while attention layers control information aggregation~\citep{attn1,attn2}, we adapt both components. Restricting adaptation to attention layers alone is insufficient for modeling high-density speech representations, making FFN adaptation essential for stable performance across diverse languages.
\section{Experiments}
\subsection{Datasets}
We combine three multilingual ASR datasets for training—FLEURS~\citep{fleurs} (968 hours, 102 languages), CommonVoice~\citep{commonvoice} (10,448 hours, 131 languages), and the Omnilingual ASR Corpus~\citep{omniasr} (2,342 hours, 348 languages)—totaling 13,758 hours across 495 unique languages. All samples were resampled to 16 kHz. As the Omnilingual ASR corpus mainly consists of long utterances, we segment them into approximately 30-second chunks using MMS-FA~\citep{mms}. The segmented OmniASR dataset is available at HuggingFace\footnote{https://huggingface.co/datasets/Sanghyang00/omniasr-molge}, and additional details on preprocessing procedures are provided in Appendix~\ref{appendix:data_prep}.

\subsection{Backbone and Baseline Models}
We evaluate the MoLGE framework using two multilingual ASR models as a backbone, which emphasize complementary aspects of speech recognition: \textbf{\textit{phonetic}} and \textbf{\textit{orthographic}} modeling. For a fair comparison, both models utilize Connectionist Temporal Classification (CTC) variants for ASR and are configured with 300M parameters. Details on the rationale behind the backbone selection are provided in Appendix~\ref{appendix:romanization}.

\noindent\textbf{LAMA-UT.}
LAMA-UT~\citep{lamaut} is a multilingual ASR framework that predicts a universal Romanized transcription in a language-agnostic manner, subsequently transliterating it into language-specific scripts using frozen LLMs. In our experiments, we use only the Romanized transcription generator to isolate the model's \textbf{\textit{phonetic}} modeling capabilities.

\noindent\textbf{Omnilingual ASR.}
Omnilingual ASR~\citep{omniasr} is a massively multilingual ASR model based on wav2vec2.0, supporting over 1600 languages. In contrast to LAMA-UT, this model directly predicts language-specific scripts, emphasizing \textbf{\textit{orthographic}} modeling.

\noindent\textbf{Baseline Configurations.}
For each backbone, we compare MoLGE against two baselines: (1) a \textit{dense model}, which uses a single ASR model without any adaptation, corresponding to the original CTC-based ASR architecture, and (2) \textit{random grouping}, where MoLGE is applied with randomly assigned language groups without any linguistic guidance.

\subsection{Training Details}
We configured the MoLGE with a fixed LoRA rank of 32 and trained it with a composite objective of CTC loss~\citep{ctc} and a routing loss, where implicit grouping uses load balancing and explicit grouping uses cross-entropy for supervised group prediction. The routing loss is weighted by 0.01 relative to the CTC loss.
All models are trained with AdamW~\citep{adamw} optimizer using a tri-stage learning rate schedule over 10k steps: linear warm-up from $5\mathrm{e}^{-6}$ to $1\mathrm{e}^{-4}$ (10\%), constant (60\%), and decay to $5\mathrm{e}^{-6}$ (30\%).

To alleviate resource and environmental imbalance in multilingual data, we adopt a multilevel sampling strategy following~\cite{mms} and~\cite{omniasr}, formulated as $p_{l,d} \propto (n_{l,d}/N)^{\beta_{L,D}}$. Here, $n_{l,d}$ denotes the number of samples for language $l$ (or corpus $d$), and $N$ is the total number of samples. We set $\beta_L = \beta_D = 0.5$. 
Additionally, we also apply data augmentation using room impulse responses (RIR)~\citep{but} and background noise~\citep{demand}, where 50\% of samples are clean, and the rest are equally split between RIR convolution and additive noise (25\% each).

\begin{table*}[!t]
\centering
\begin{threeparttable}
    \begin{tablenotes}[flushleft]
        \scriptsize
        \item ~~~\textbf{Low-Resource}: $<$10h, \textbf{Mid-Resource}: 10--50h, \textbf{High-Resource}: $>$50h
    \end{tablenotes}

    \resizebox{0.97\textwidth}{!}{%
    \begin{tabular}{c|c|c|cc|cc|cc|cc} \toprule
    \multirow{2}{*}{\textbf{Backbone}} & \multirow{2}{*}{\textbf{Method}} & \multirow{2}{*}{\makecell{\textbf{Activated}\\\textbf{Params.}}} & \multicolumn{2}{c|}{\textbf{Average}} & \multicolumn{2}{c}{\textbf{Low-Resource}} & \multicolumn{2}{c}{\textbf{Mid-Resource}} & \multicolumn{2}{c}{\textbf{High-Resource}} \\ 
    & &  & \textbf{CER ($\downarrow$)} & \textbf{Acc. ($\uparrow$)} & \textbf{CER ($\downarrow$)} & \textbf{Acc. ($\uparrow$)} & \textbf{CER ($\downarrow$)} & \textbf{Acc. ($\uparrow$)} & \textbf{CER ($\downarrow$)} & \textbf{Acc. ($\uparrow$)} \\ 
    \midrule\midrule
    
    \multirow{3}{*}{\makecell{LAMA-UT \\ (Phonetic)}} & Dense & 300 M & 25.15 & - & 26.16 & - & 19.06 & - & 21.84 & - \\
    & Random & 343 M & 24.31 & 68.44 & 25.46 & 69.88 & 17.66 & 77.77 & 19.93 & 23.84 \\ \cmidrule{2-11}
    & \textbf{MoLGE} & \textbf{343 M} & \textbf{22.27} & \textbf{93.96} & \textbf{23.43} & \textbf{94.84} & \textbf{15.55} & \textbf{93.43} & \textbf{17.99} & \textbf{80.51} \\ 
    
    \midrule\midrule
    
    \multirow{3}{*}{\makecell{OmniASR \\ (Orthographic)}} & Dense & 300 M & 29.30 & - & 31.41 & - & 20.68 & - & 13.38 & - \\ 
    & Random & 353 M & 25.81 & 53.24 & 27.59 & 53.32 & 18.12 & 67.84 & 13.37 & 19.37 \\ \cmidrule{2-11}
    & \textbf{MoLGE} & \textbf{353 M} & \textbf{23.73} & \textbf{94.44} & \textbf{25.44} & \textbf{95.27} & \textbf{16.34} & \textbf{89.35} & \textbf{11.72} & \textbf{91.91} \\
    
    \bottomrule
    \end{tabular}%
    }

    \caption{Performance on LAMA-UT and Omnilingual ASR (OmniASR) models. We report CER (\%) and routing accuracy (Acc., \%) breakdown by resource availability.}
    \label{tab:main}
\end{threeparttable}
\vspace{-10pt}
\end{table*}
\section{Results and Analysis}
\begin{figure*}[!t]
  \centering
  \includegraphics[width=0.95\textwidth]{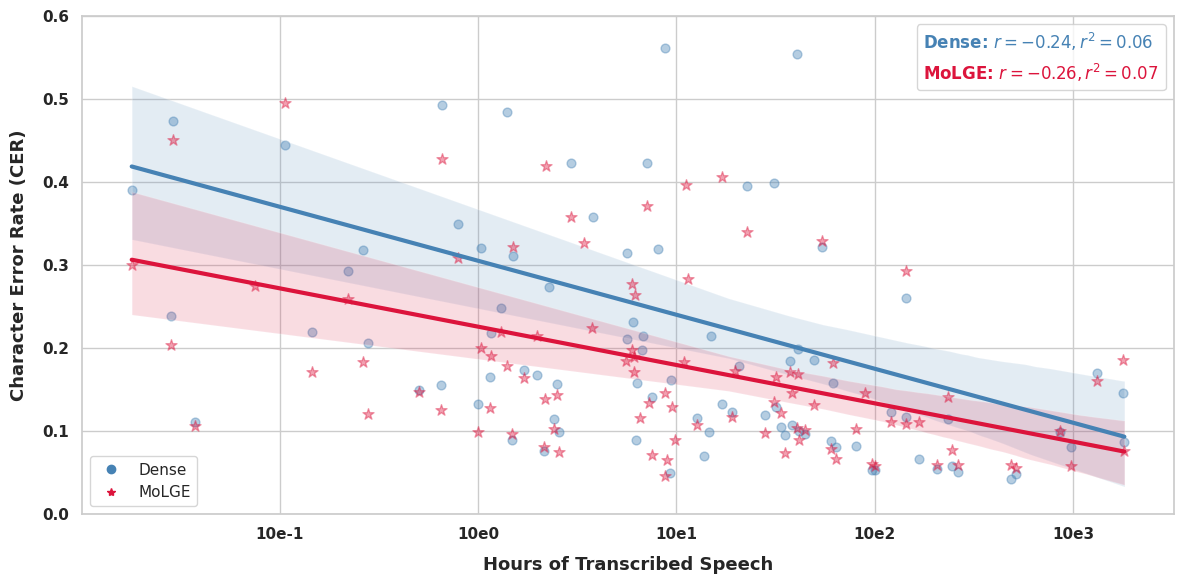}
  \caption{Regression plot between CER and training data. Each point indicates the mean CER per language, and the shaded region represents the confidence interval.}
  \label{fig:unified_regression}
\end{figure*}
\subsection{Effects of MoLGE on Massively Multilingual ASR}
\label{sec:domain}
\noindent\textbf{Performance Across Linguistic Domains.}
As shown in Table~\ref{tab:main}, applying MoLGE yields substantial improvements in both phonetic and orthographic modeling for speech recognition, with relative gains of 11.5\% and 19.0\% in CER, respectively. 
Interestingly, improvements in the phonetic domain are relatively modest compared to the orthographic domain, and mid-resource languages sometimes exhibit lower CER than high-resource languages. This suggests that performance in the phonetic domain (i.e., a simplified latent space~\citep{roman1,roman2,roman3}) is influenced not only by data scale but also by factors such as representation complexity.
In contrast, in the orthographic domain, MoLGE significantly improves both CER and routing accuracy compared to dense and randomly grouped baselines, highlighting the benefits of incorporating language-aware structure in more complex linguistic spaces.
Overall, these results indicate that the effectiveness of MoLGE is closely tied to linguistic complexity, with linguistic priors becoming increasingly beneficial for modeling richer orthographic and language-specific structures.

\noindent\textbf{Data–Performance Correlation and Generalization Ability.}
In Figure~\ref{fig:unified_regression}, we illustrate the regression plot, which visualizes the correlation between CER and training data in the orthographic domain (i.e., OmniASR).
When applying MoLGE, the correlation between training data size and CER becomes marginally stronger, with the correlation coefficient $r$ decreasing from -0.24 to -0.26, leading to more predictable learning dynamics. 
Moreover, the regression slope becomes less steep, with CER on low-resource languages dropping substantially from 42\% to 30\%, demonstrating that MoLGE narrows the performance gap across resource levels.
Furthermore, confidence intervals become notably narrower, indicating that languages with similar training data now achieve more consistent performance. 
Taken together, these results demonstrate that MoLGE enhances robustness in low-resource languages and enables reliable generalization across languages of varying resource levels.

\begin{figure*}[t]
    \centering
    \begin{subfigure}{0.24\textwidth}
        \centering
        \includegraphics[width=\linewidth]{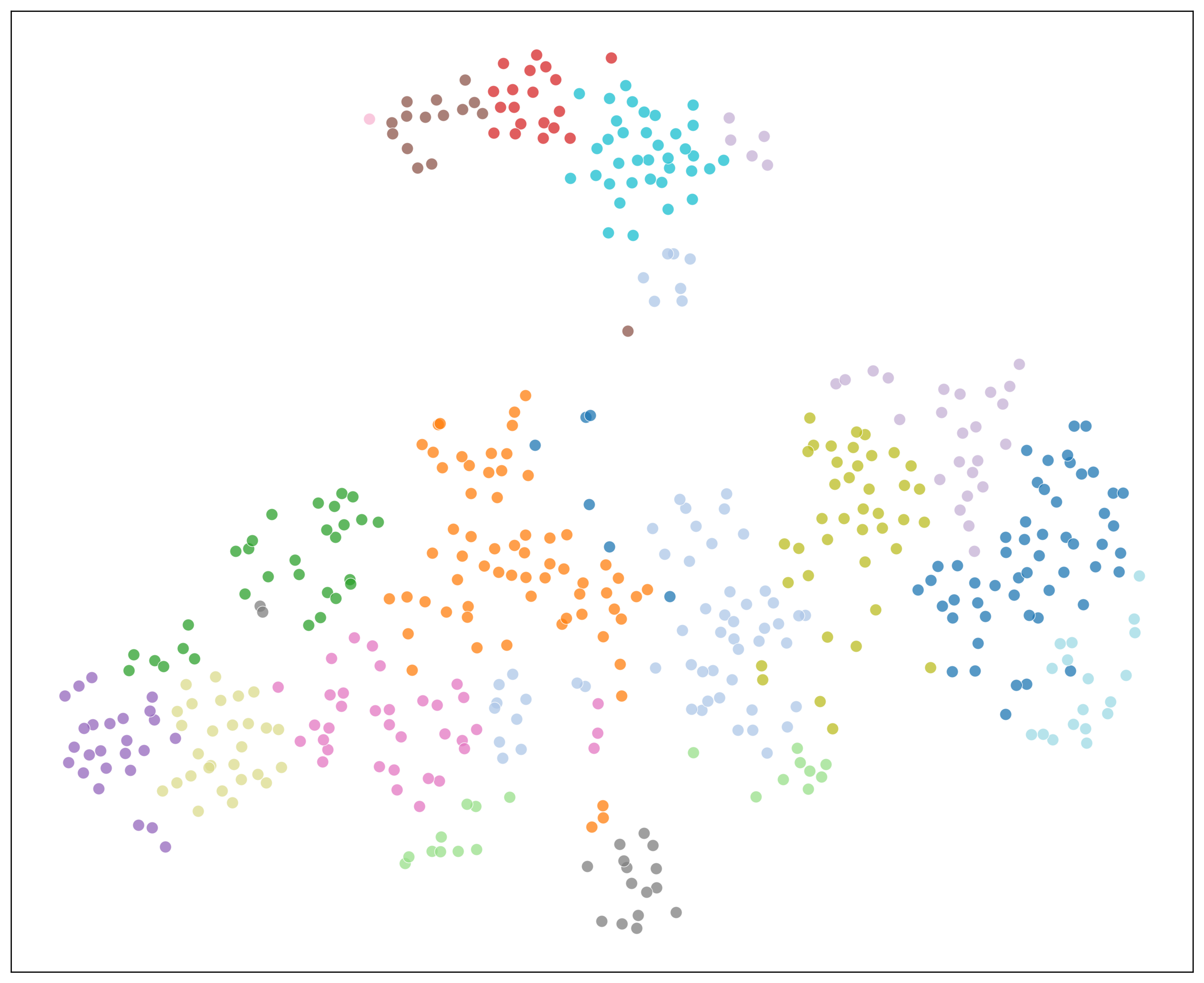}
        \caption{SSL}
        \label{fig:sub1}
    \end{subfigure}
    \hfill
    \begin{subfigure}{0.24\textwidth}
        \centering
        \includegraphics[width=\linewidth]{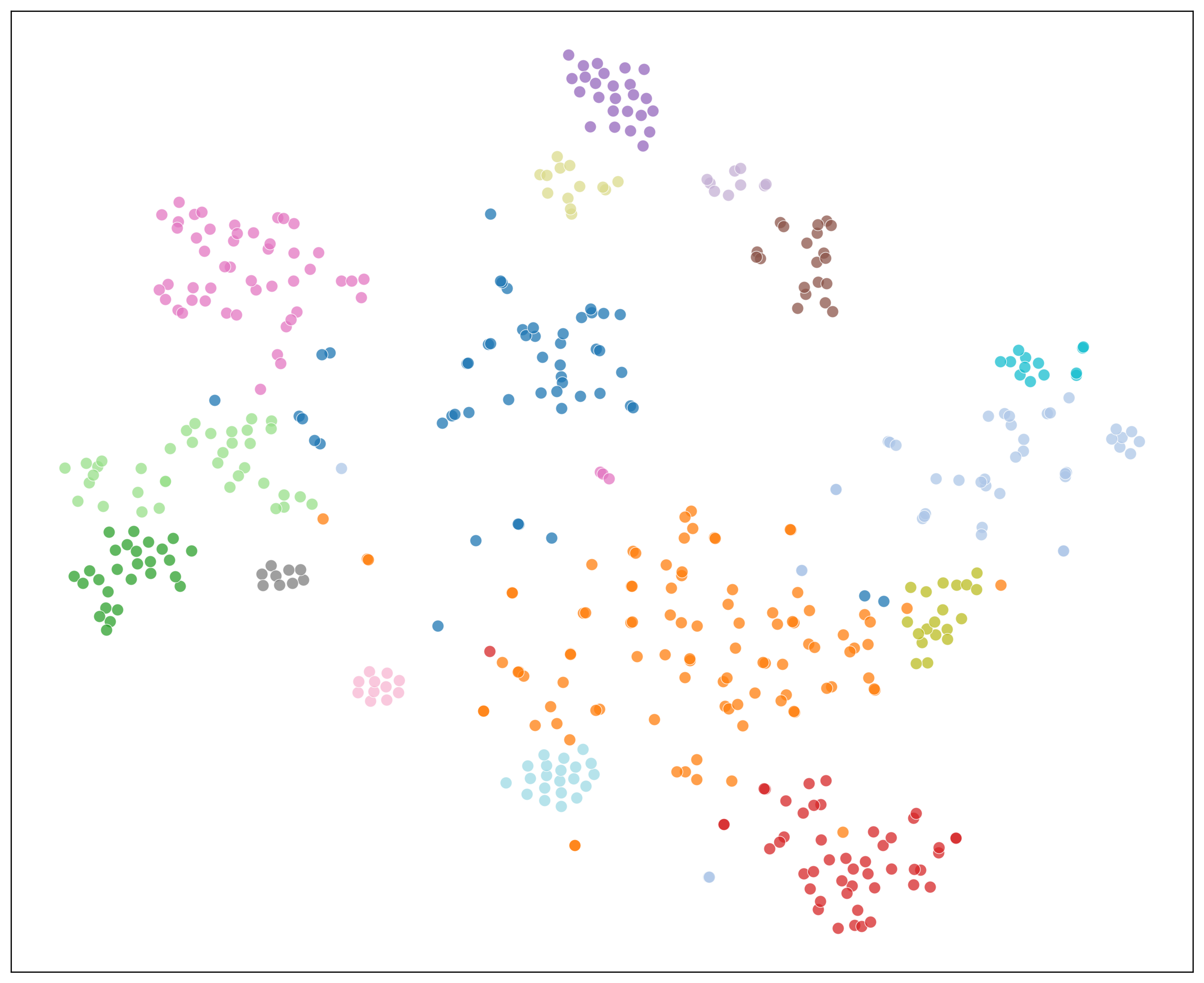}
        \caption{LID}
        \label{fig:sub2}
    \end{subfigure}
    \hfill
    \begin{subfigure}{0.24\textwidth}
        \centering
        \includegraphics[width=\linewidth]{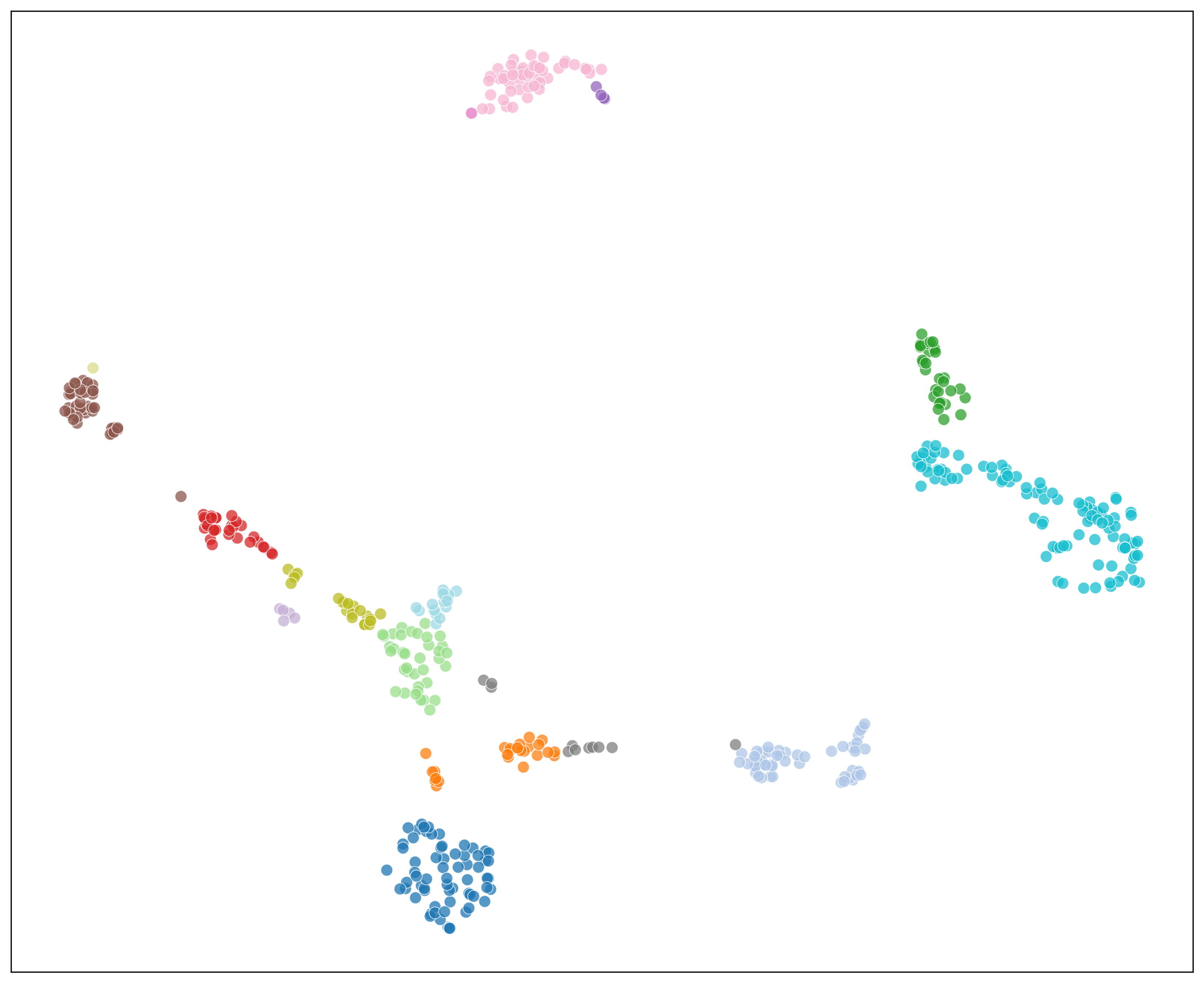}
        \caption{Geographic}
        \label{fig:sub3}
    \end{subfigure}
    \hfill
    \begin{subfigure}{0.24\textwidth}
        \centering
        \includegraphics[width=\linewidth]{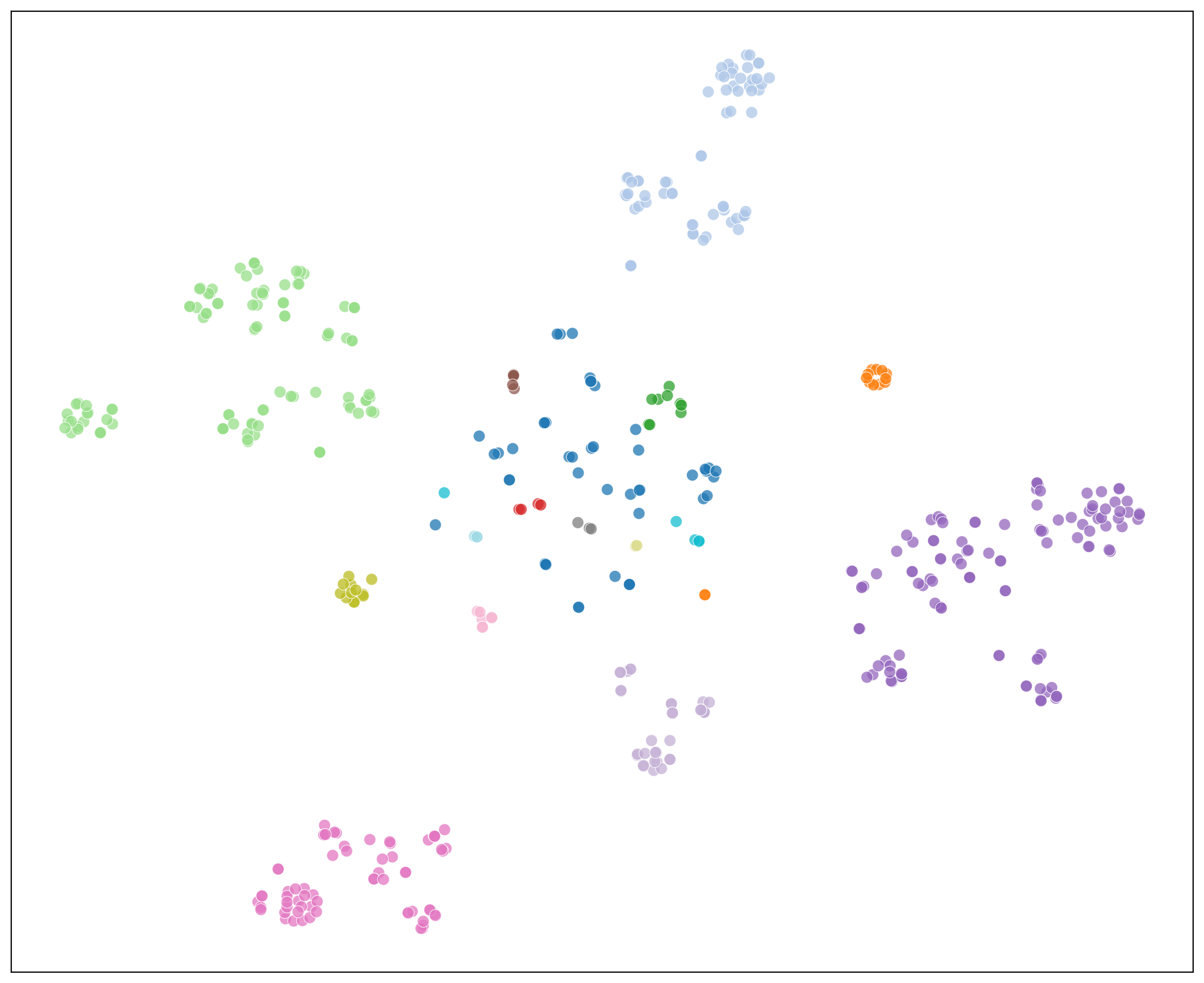}
        \caption{Genetic}
        \label{fig:sub4}
    \end{subfigure}
    
    \caption{Visualization of language clustering across different grouping strategies. Each point represents a language, and colors indicate cluster assignments in the t-SNE space.}
    \label{fig:cluster}
\end{figure*}
\subsection{Analysis on Cluster Formation and Language Grouping}
\noindent\textbf{Cluster Formation Across Guidance Methods.}
Figure~\ref{fig:cluster} illustrates clustering results using t-SNE across four explicit grouping strategies. Embedding-guided methods generally produce weakly separated clusters, with SSL representations showing particularly fragmented structures. While LID-based clustering yields clearer boundaries than SSL, it remains more dispersed than knowledge-guided methods, likely due to the high-dimensional nature of embeddings (e.g., a feature dimensionality of 1,280 for both SSL and LID models).

In contrast, knowledge-guided methods produce more structured and well-separated clusters than embedding-guided approaches, owing to the compact nature of knowledge vectors. Geographic grouping forms coarse but coherent clusters, while genetic grouping leaves some languages scattered near the center.
We attribute this behavior to the sensitivity of knowledge-guided methods to the imputation mechanism of URIEL+. Since URIEL+ imputes missing features based on information from related or higher-level language families, genetic grouping often assigns linguistically diverse languages—including isolates such as Japanese (jpn) and Korean (kor), dialects such as Liberian English (lir), and Maria (mrr), and indigenous languages such as Hunjara-Kaina Ke (hkk) and Ömie (aom)—to the same group. Additional visualizations with language annotations are provided in Appendix~\ref{appendix:cluster}.

\begin{table}[!t]
    \centering
    \small
    \begin{minipage}{0.7\linewidth}
        \centering
        \resizebox{\textwidth}{!}{%
        \begin{tabular}{c|c|cc|c|c|cc}
            \toprule
            \multicolumn{4}{c|}{\textbf{LAMA-UT (Phonetic)}} & \multicolumn{4}{c}{\textbf{OmniASR (Orthographic)}} \\
            \textbf{Guidance} & \textbf{Method} & \textbf{CER  ($\downarrow$)} & \textbf{Acc.  ($\uparrow$)} & \textbf{Guidance} & \textbf{Method} & \textbf{CER  ($\downarrow$)} & \textbf{Acc.  ($\uparrow$)} \\ \midrule\midrule
            \multirow{2}{*}{Implicit}  & Frame      & 22.30 & -     & \multirow{2}{*}{Implicit}  & Frame      & 24.30 & -     \\
                                       & Sample     & 22.32 & -     &                            & Sample     & 24.51 & -     \\ \cmidrule{1-8}
            \multirow{2}{*}{Embedding} & SSL        & \underline{22.29} & 72.17 & \multirow{2}{*}{Embedding} & SSL        & 23.82 & 64.16 \\
                                       & \textbf{LID} & \textbf{22.27} & \textbf{93.96} &                    & \textbf{LID} & \underline{23.74} & \textbf{94.70} \\ \cmidrule{1-8}
            \multirow{2}{*}{Knowledge} & \textbf{Geographic} & \underline{22.29} & \underline{93.51} & \multirow{2}{*}{Knowledge} & \textbf{Geographic} & \textbf{23.73} & \underline{94.44} \\
                                       & Genetic    & \underline{22.29} & 92.24 &                            & Genetic    & 23.80 & 90.79 \\ \bottomrule
        \end{tabular}%
        }
        \caption{Average transcription performance (CER) and routing accuracy (Acc.) on six guidance methods, respectively.}
        \label{tab:molge_summary}
    \end{minipage}
    \hfill
    \begin{minipage}{0.27\linewidth}
        \centering
        \begin{threeparttable}
            \begin{tablenotes}[flushleft]
                \scriptsize
                \item \textbf{S.E.}: Shared Expert
                \item \textbf{H.A.}: Hierarchical Adapt.
            \end{tablenotes}
            \vspace{2pt}
            \resizebox{\textwidth}{!}{%
            \begin{tabular}{l|cc} 
            \toprule
            \textbf{Model} & \textbf{CER ($\downarrow$)} & \textbf{Acc. ($\uparrow$)} \\ 
            \midrule\midrule
            \textbf{MoLGE}           & \textbf{23.73} & \textbf{94.44} \\ \midrule
            $-$ S. E.        & 24.70 & 92.77 \\ 
            $-$ H. A. (M)   &   23.85  &  73.12 \\
            $-$ H. A. (L)        & 27.24 & - \\ 
            \bottomrule
            \end{tabular}%
            }
            \caption{Ablation study on architecture design.}
            \label{tab:molge_variants}
        \end{threeparttable}
    \end{minipage}
\end{table}
\noindent\textbf{Effect of Cluster Coherence.}
Table~\ref{tab:molge_summary} presents the performance of MoLGE under various grouping configurations.
Across both phonetic and orthographic spaces, explicit guidance consistently outperforms implicit guidance, supporting our hypothesis that prior-based grouping is more effective than router-driven grouping.

To further analyze this trend, we break down the results by representation space.
In the phonetic space, performance differences across grouping strategies are minimal, also suggesting that language priors play a limited role in a simplified phonetic latent space. 
In contrast, in the orthographic space, incorporating language priors consistently improves performance, and more coherent grouping leads to lower CER, as evidenced by the superior performance of LID over SSL embeddings and geographic over genetic vectors.

Overall, these findings indicate that the effectiveness of language grouping arises from the interaction between linguistic complexity, the information density of linguistic priors, and cluster coherence. As linguistic complexity increases, richer priors yield more coherent groupings, which in turn translate into improved performance.

\begin{table}[!t]
    \centering
    \small

    \begin{minipage}{0.65\linewidth}
        \centering
        \resizebox{\textwidth}{!}{%
        \begin{tabular}{lrr|lrr}
        \toprule
        \multicolumn{3}{c|}{\textbf{Indic Languages}} &
        \multicolumn{3}{c}{\textbf{Nigerian Languages}} \\
        \cmidrule(r){1-3} \cmidrule(l){4-6}
        \textbf{Language} & \textbf{Hours} & \textbf{CER ($\downarrow$)} &
        \textbf{Language} & \textbf{Hours} & \textbf{CER ($\downarrow$)} \\
        \midrule\midrule
        Tamil              & 89.13 & 14.73 & Hausa           & 12.24 & 12.58 \\
        Bengali            & 40.38 & 10.68 & Yoruba          & 10.82 & 22.46 \\
        Marathi            & 11.44 & 9.90  & Tangale         & 7.01  & 20.15 \\
        Chitwania Tharu    & 5.43  & 10.70 & Burak           & 6.70  & 19.82 \\
        Odia               & 5.05  & 19.01 & Tula            & 5.68  & 21.63 \\
        Santali (Ol Chiki) & 0.07  & 21.21 & Cakfem-Mushere  & 5.53  & 24.00 \\
        \bottomrule
        \end{tabular}
        }
        \caption{ASR performance of MoLGE on six Indic and Nigerian languages, respectively.}
        \label{tab:knowledge_transfer}
    \end{minipage}
    \hfill
    \begin{minipage}{0.32\linewidth}
        \centering
        \resizebox{\textwidth}{!}{%
        \begin{tabular}{c|c|c}
        \toprule
        \textbf{\# Experts} & \textbf{Total Params.} & \textbf{CER ($\downarrow$)} \\
        \midrule\midrule
        8  & 409M & 27.34 \\
        16 & 465M & \textbf{24.51} \\
        32 & 580M & 24.52 \\
        \bottomrule
        \end{tabular}
        }
        \caption{Effect of the number of experts in sample-level implicit grouping.}
        \label{tab:num_experts}
    \end{minipage}
\end{table}
\subsection{Knowledge Transfer Across Resource Levels}
Table~\ref{tab:knowledge_transfer} presents ASR performance on representative mid- and low-resource languages from India and Nigeria. To investigate whether MoLGE promotes knowledge sharing across resource levels, we compare training hours and recognition performance within each language group.

As shown in Table~\ref{tab:knowledge_transfer}, languages belonging to the same linguistic family are commonly assigned to the same group and exhibit relatively consistent CER trends despite large differences in training hours. For example, several low-resource languages achieve CERs comparable to those of substantially higher-resource languages within the same group. Similar patterns are observed across both the Indic and Nigerian language groups.
Overall, these findings suggest that MoLGE facilitates knowledge transfer within each language group, helping reduce performance disparities between high- and low-resource languages.

\subsection{Ablation Study}
\noindent\textbf{Architectural Design.}
In Table~\ref{tab:molge_variants}, we conduct an ablation study by sequentially removing key components, starting with the shared experts in the upper Transformer layer MoLE module, followed by the speech-aware hierarchical adaptation.
Removing the shared experts degrades both CER and routing accuracy, highlighting their importance in effective acoustic modeling within the MoLE framework.

Furthermore, eliminating the hierarchical adaptation layer consistently deteriorated model performance. Replacing MoLGE with pure MoLE (indicated by 'M') slightly impaired performance despite higher parameter counts, while replacing it with vanilla LoRA (indicated by 'L') led to a severe performance drop. This confirms the necessity of hierarchical adaptation for integrating language-specific acoustic and textual features.

\noindent\textbf{Number of Experts.}
In addition to the theoretical motivation discussed in Section~\ref{sec:langgroup}, we conducted a pilot study to examine the impact of the number of experts. As shown in Table~\ref{tab:num_experts}, reducing the number of experts from 16 to 8 resulted in a substantial degradation in recognition performance. Increasing the number of experts to 32, however, provided negligible improvement (CER from 24.51 to 24.52) while increasing the model size from 465M to 580M parameters. Based on these empirical and theoretical bases, we adopted 16 experts, which offers the best trade-off between recognition performance and computational efficiency.

\section{Conclusion}
In this paper, we introduced Mixture of Language Group Experts (MoLGE), an efficient and scalable framework for massively multilingual automatic speech recognition. We show that incorporating architectural inductive biases on linguistic structure, together with language grouping, improves transcription performance while enabling scalable modeling across a large number of languages and mitigating the curse of multilinguality.
Specifically, our approach integrates speech-oriented parameter-efficient finetuning with structured language grouping, providing a unified framework for scalable massively multilingual speech recognition. We further conduct a comprehensive analysis of diverse grouping strategies, demonstrating their effectiveness across varying resource conditions and their role in promoting expert specialization.
Finally, experiments across phonetic and orthographic domains reveal that the benefits of language-aware modeling become more pronounced as linguistic complexity increases, with explicit language priors providing more coherent and effective guidance. 
We hope this work serves as a foundation for scaling multilingual speech systems and inspires future research on structured, language-aware modeling.
\section*{Acknowledgment}
This work was supported by the National Research Foundation of Korea (NRF) grant funded by the Korea government Ministry of Science and ICT (MSIT) (RS-2026-25468664).

\bibliography{colm2026_conference}
\bibliographystyle{colm2026_conference}

\newpage
\appendix
\begin{figure*}[t]
    \centering
    \begin{subfigure}{0.48\textwidth}
        \centering
        \includegraphics[width=\linewidth]{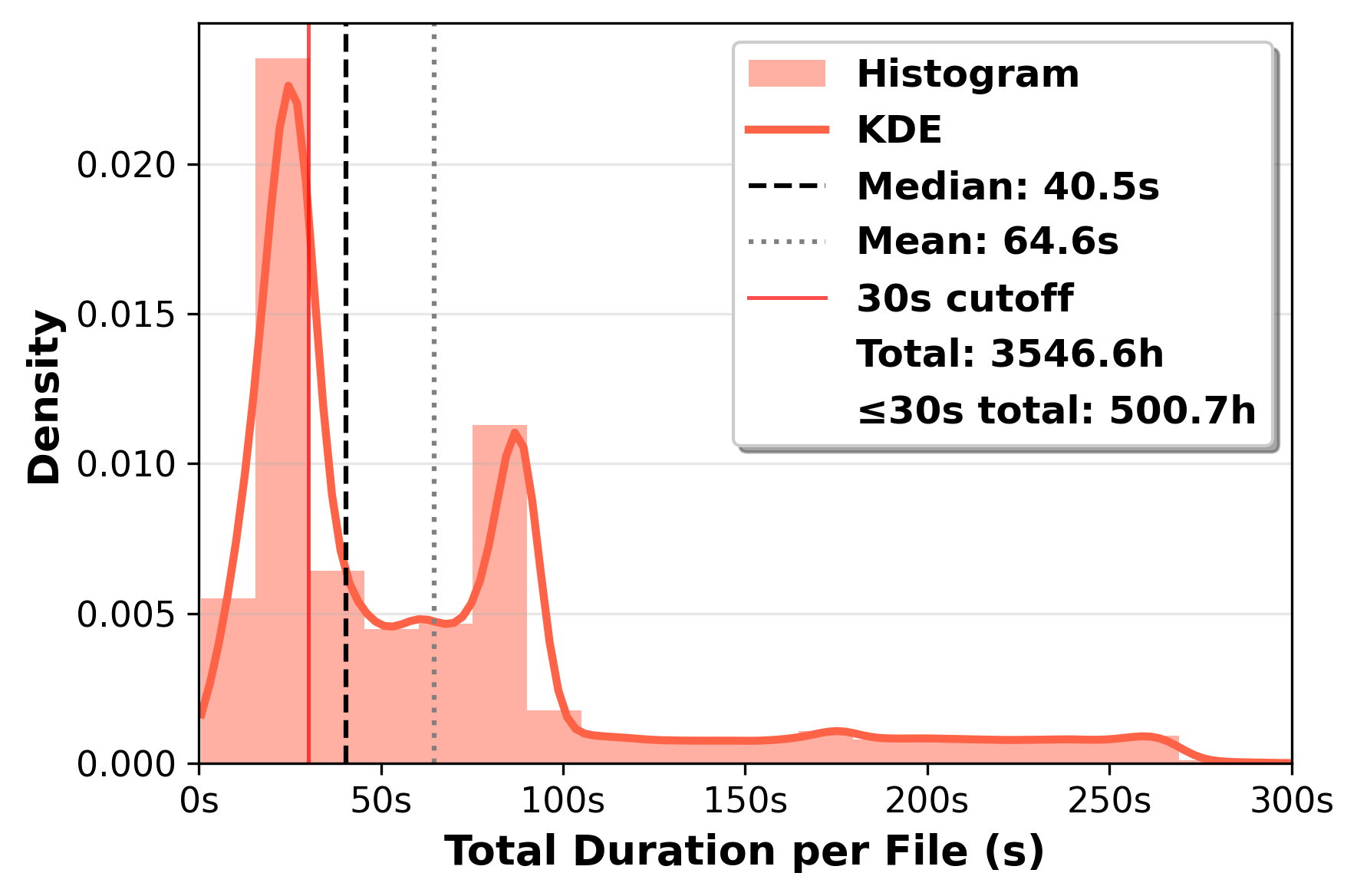}
        \caption{Duration of the original dataset.}
        \label{fig:pre_data}
    \end{subfigure}
    \hfill 
    \begin{subfigure}{0.48\textwidth}
        \centering
        \includegraphics[width=\linewidth]{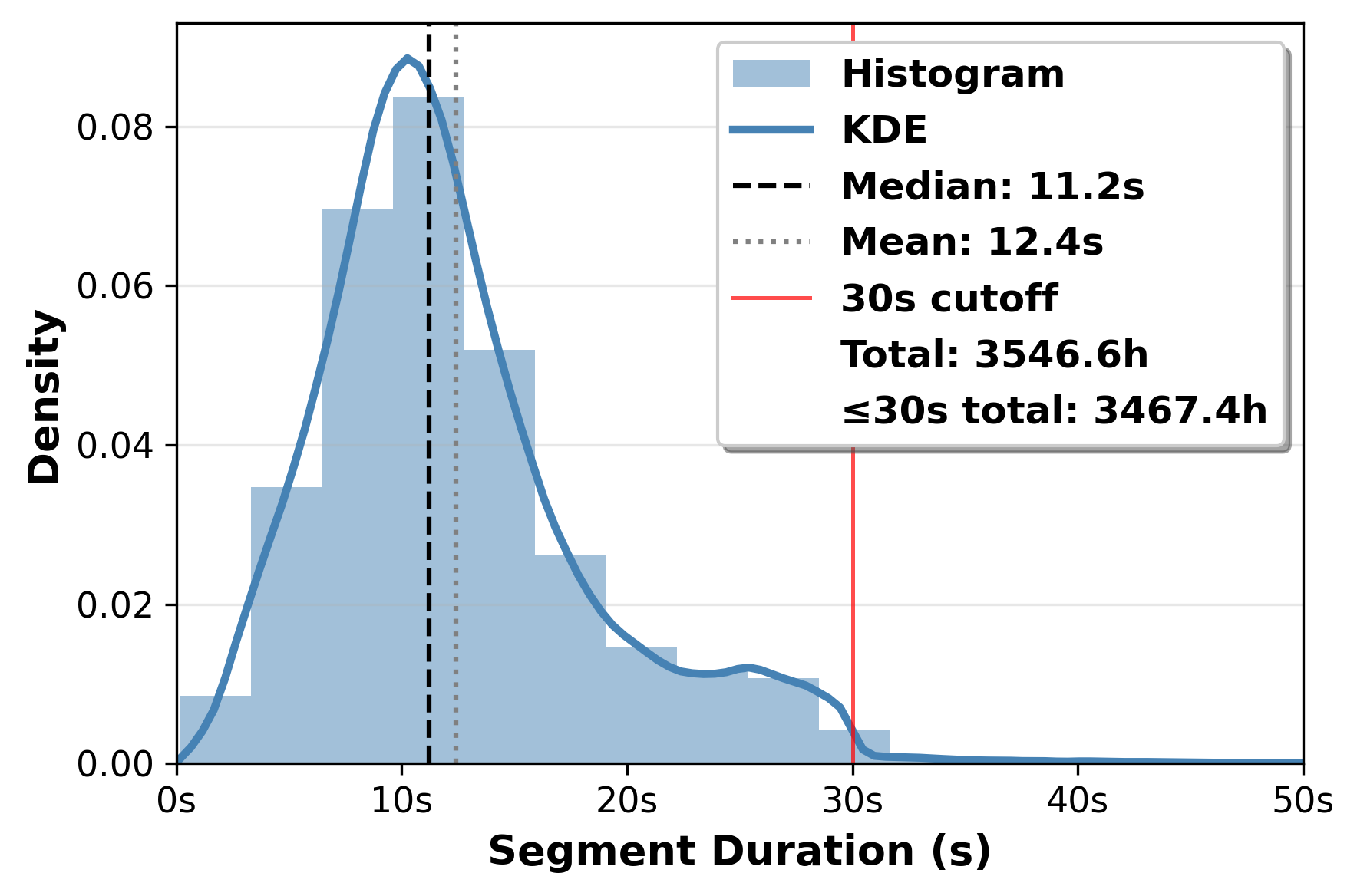}
        \caption{Duration of the dataset after segmentation.}
        \label{fig:post_data}
    \end{subfigure}
    
    \caption{Duration distributions in the omnilingual ASR corpus, aggregated across all splits.}
    \label{fig:app_data_dist}
\end{figure*}
\section{Dataset Details and Preprocessing Strategy}
\label{appendix:data_prep}
\subsection{Dataset Details}
\noindent\textbf{FLEURS.}
FLEURS is a multilingual dataset encompassing 102 languages, specifically curated for ASR and speech translation. 
The corpus consists of read speech from Wikipedia articles, providing relatively clean recordings. Each language contains approximately 10 hours of supervised data, offering a balanced foundation for high-quality acoustic modeling.

\noindent\textbf{CommonVoice.}
CommonVoice is a large-scale crowdsourced dataset covering 131 languages, exhibiting substantial acoustic variability, including diverse microphone qualities and background noise levels. We use the official splits from CommonVoice 22.0.

\noindent\textbf{Omnilingual ASR Corpus.}
Omnilingual ASR corpus is a massively multilingual dataset spanning 348 languages, with a significant emphasis on low-resource languages and long-form speech. To mitigate training instability associated with long utterances, we perform audio segmentation via forced alignment.

\subsection{Preprocessing Details of Omnilingual ASR Corpus}
\noindent\textbf{Dataset Characteristics.}
The Omnilingual ASR corpus differs from conventional ASR datasets in that it captures naturally occurring speech by prompting speakers with over 1,500 queries and recording their responses.
As a result, it contains many long utterances exceeding 30 seconds, as shown in Figure~\ref{fig:app_data_dist}. Rather than discarding them, which would significantly reduce the data, we apply a segmentation strategy to preserve the data capacity.

\noindent\textbf{Segmentation Strategy.}
To address the aforementioned issue, we adopt a three-stage segmentation pipeline for utterances longer than 30 seconds: (1) segmentation using hard punctuation (e.g., periods, question marks, exclamation marks), (2) segmentation using soft punctuation (e.g., commas), and (3) forced segmentation based on a character limit. While most utterances can be segmented within 30 seconds using punctuation, we observe that some languages—particularly African languages—often contain long, punctuation-free running sentences. To address this, we apply forced segmentation at approximately 150 characters, splitting at the nearest word boundary.
All segmentation is performed using MMS-FA. After segmentation, we filter samples based on alignment scores on a per-language basis, pruning low-quality segments to ensure overall data quality.

\section{Utilization of Romanization as a Phonetic Proxy}
\label{appendix:romanization}
\subsection{Representation Analysis in Multilingual ASR}
Recent multilingual speech benchmarks, such as ML-SUPERB~\citep{mlsuperb,mlsuperb2}, have demonstrated the effectiveness of self-supervised speech models for multilingual ASR across diverse languages. Most existing studies, however, primarily focus on improving downstream recognition performance through larger pretrained models, scaling strategies, or multilingual evaluation benchmarks.

In contrast, relatively little attention has been paid to how different transcription representations influence multilingual speech modeling. In particular, it remains unclear whether language-aware modeling techniques exhibit different behaviors when operating on orthographic or phonetic representations. Since our goal is not to introduce a new ASR backbone but to understand the effectiveness of language-aware expert grouping, we intentionally analyze MoLGE under two complementary representation spaces.

\subsection{Romanization as a Scalable Alternative}
Ideally, phonetic representations would be constructed using language-specific phoneme sequences, such as IPA-based transcriptions. In this context, we also considered adopting multilingual phonetic ASR benchmarks, such as ZIPA~\citep{zipa} and PRiSM~\citep{prism}. However, constructing reliable phonetic transcriptions at the scale considered in this work (495 languages) remains impractical.

The primary challenge lies in multilingual grapheme-to-phoneme (G2P) conversion. Existing multilingual G2P toolkits provide only limited language coverage. For example, Phonemizer~\citep{phonemizer} and CharsiuG2P~\citep{charsiug2p} support approximately 120 and 100 languages, respectively, while broader-coverage systems such as Transphone~\citep{transphone} rely extensively on zero-shot G2P generation, making phonetic quality difficult to guarantee for many low-resource languages. Consequently, generating consistent, high-quality phoneme transcriptions across hundreds of languages is currently infeasible.

Romanization provides a practical alternative that can be consistently applied across a substantially broader range of languages while preserving meaningful phonetic information. Consequently, it has been adopted in several large-scale multilingual speech systems, including MMS, where Romanized transcriptions are used for nearly one thousand languages. Although Romanization is not equivalent to true phonetic representations, prior studies have demonstrated that it serves as an effective phonetic proxy for multilingual speech modeling.

\section{Code-Switching and MoLGE}
To further investigate the linguistic behavior of MoLGE under code-switching scenarios, we analyze two public corpora with time-aligned language annotations: DISPLACE 2024~\citep{displace} and the South African Soap Opera corpus~\citep{sasoapopera}
The analysis indicates that, under sentence-level routing, the assigned expert group is largely determined by the matrix language, with agreement rates of 78.29\% and 69.91\% on DISPLACE 2024 and the South African Soap Opera corpus, respectively. In contrast, frame-level routing does not exhibit a clear or consistent language-dependent grouping pattern.

This behavior is likely attributable to the receptive field of the S3M backbone, where each frame represents approximately 25ms of acoustic context. Such short temporal windows are primarily optimized for capturing local phonetic information, making it difficult to reliably infer language identity at the frame level, particularly in code-switching speech.

\section{Detailed Visualization on Language Clusters}
\label{appendix:cluster}
\begin{figure*}[!t]
    \centering
    
    \begin{subfigure}{0.92\textwidth}
        \centering
        \includegraphics[width=\linewidth]{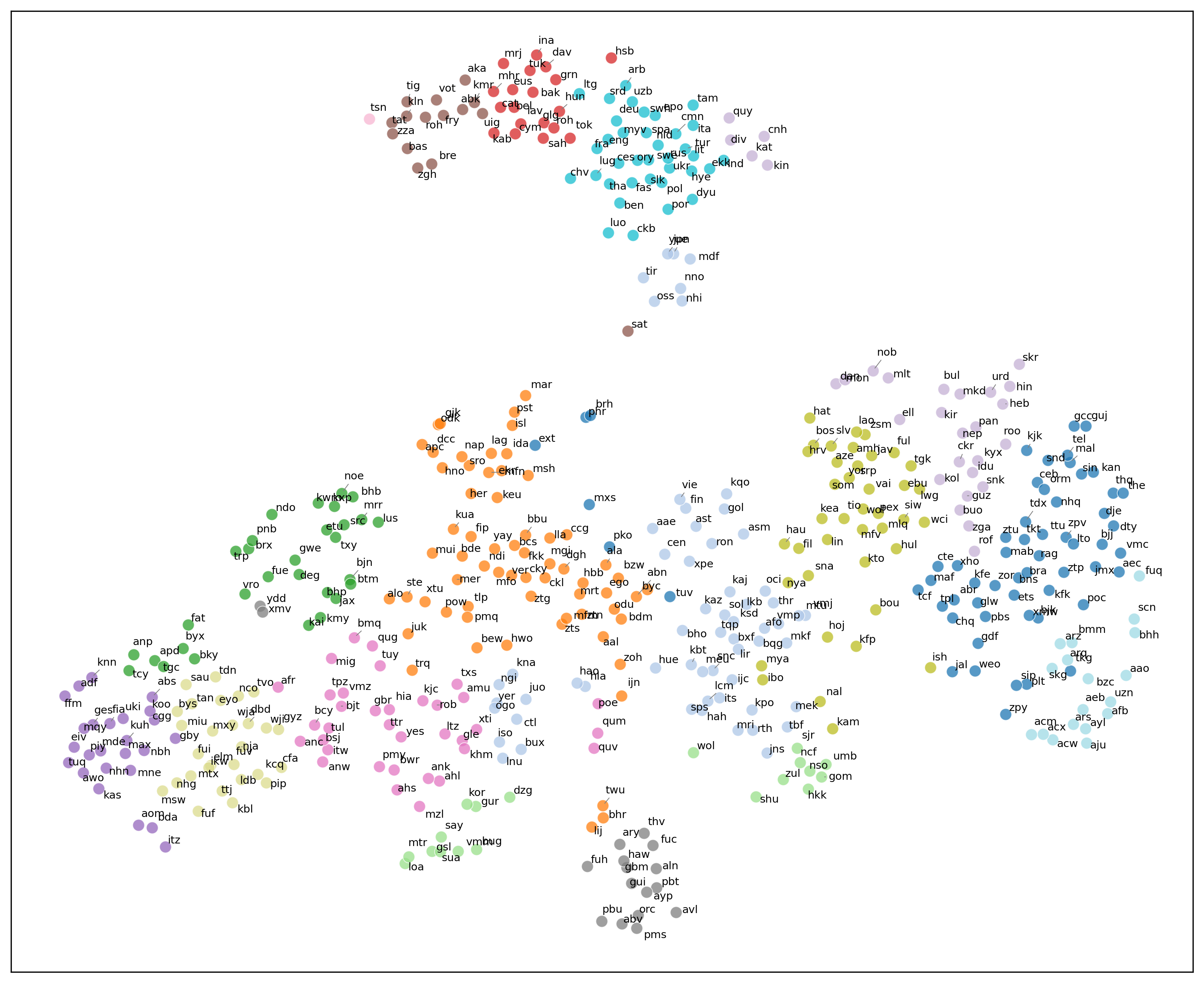}
        \caption{Cluster formation using the SSL embedding.}
        \label{fig:app_geographic}
    \end{subfigure}

    \begin{subfigure}{0.92\textwidth}
        \centering
        \includegraphics[width=\linewidth]{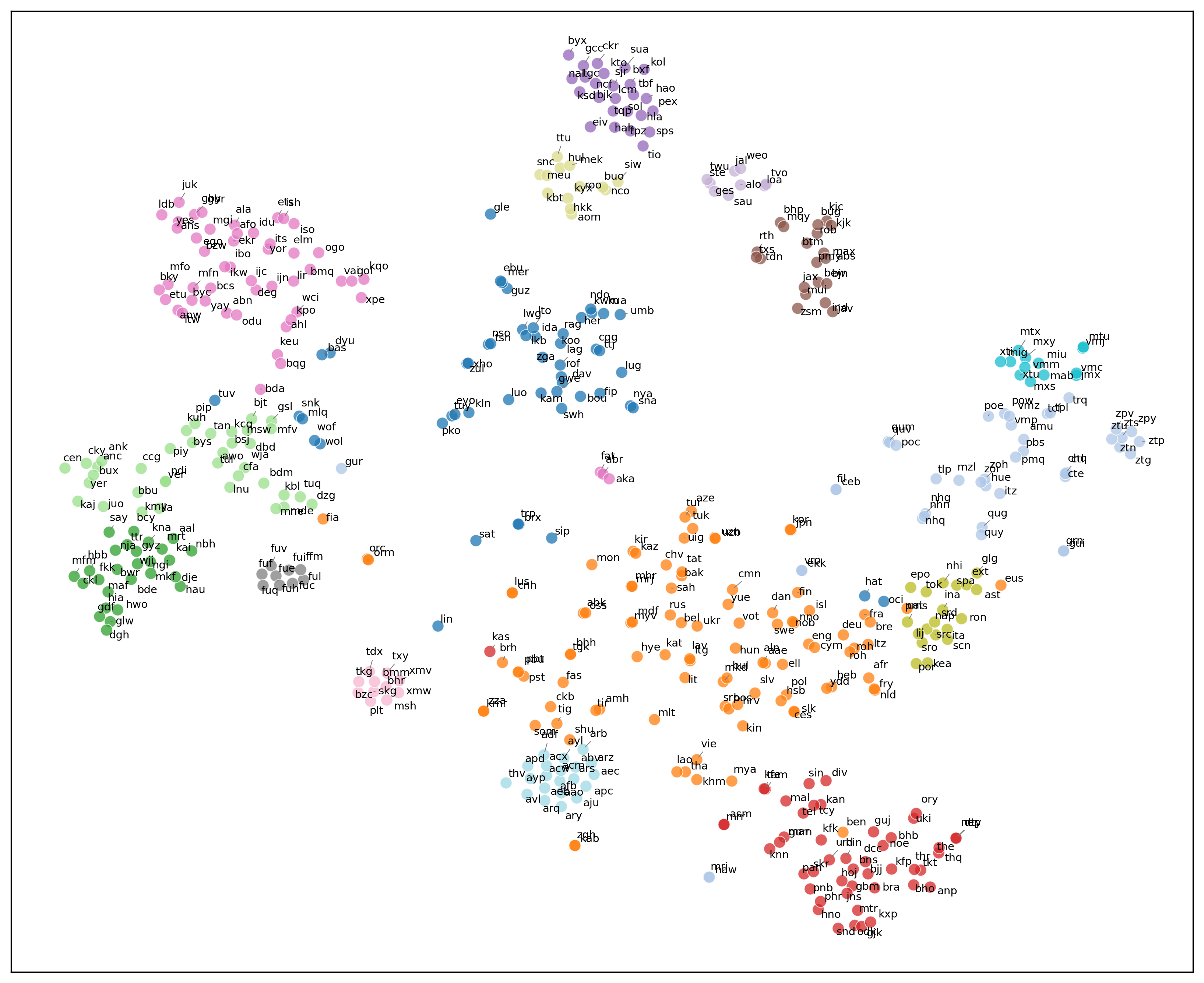}
        \caption{Cluster formation using the LID embedding.}
        \label{fig:app_genetic}
    \end{subfigure}
    
    \caption{Comparison of embedding-guided clusters.}
    \label{fig:app_clusters_comparison_embedding}
\end{figure*}
\begin{figure*}[!t]
    \centering
    
    \begin{subfigure}{0.92\textwidth}
        \centering
        \includegraphics[width=\linewidth]{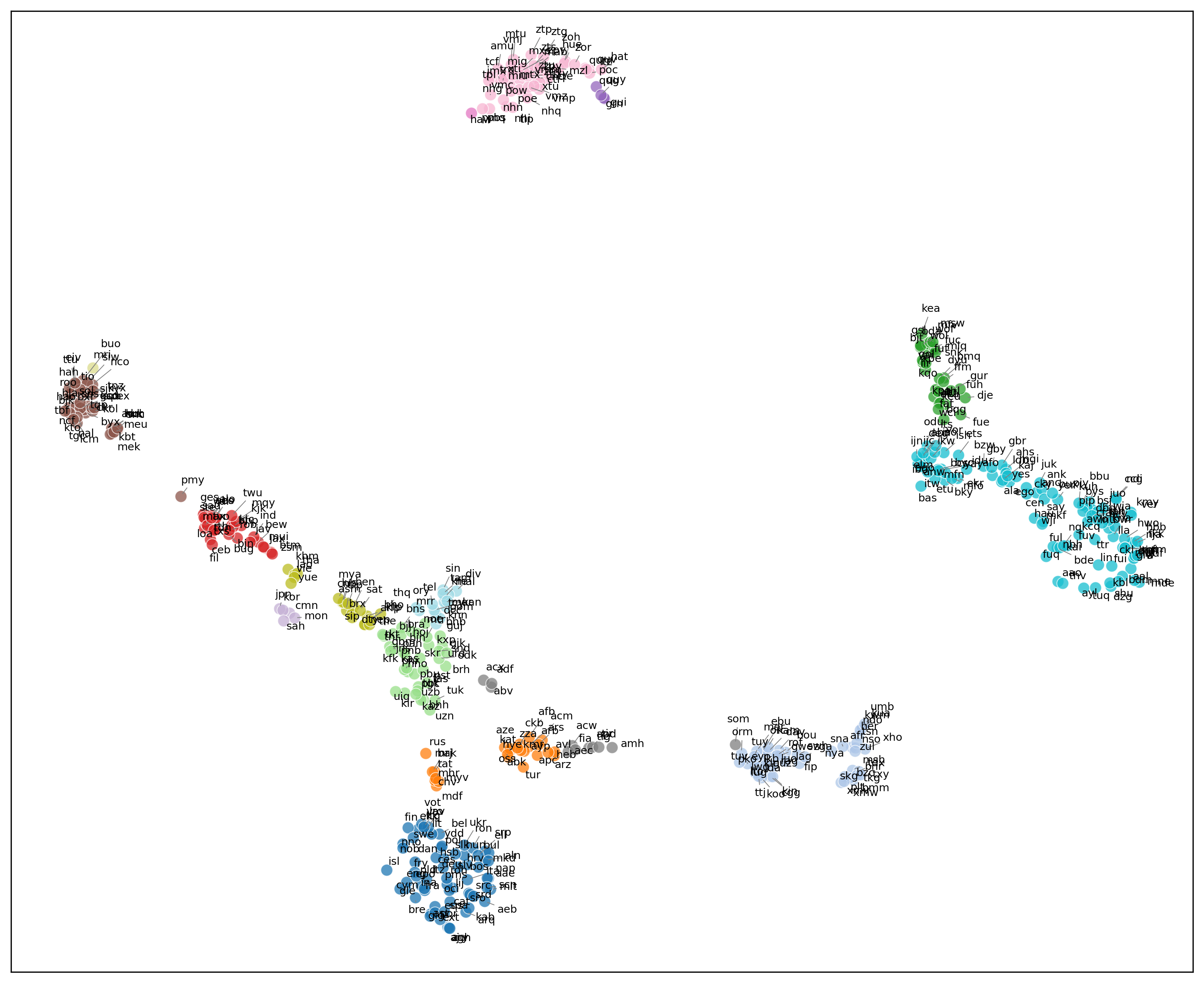}
        \caption{Cluster formation using the geographic vector.}
        \label{fig:app_geographic}
    \end{subfigure}

    \begin{subfigure}{0.92\textwidth}
        \centering
        \includegraphics[width=\linewidth]{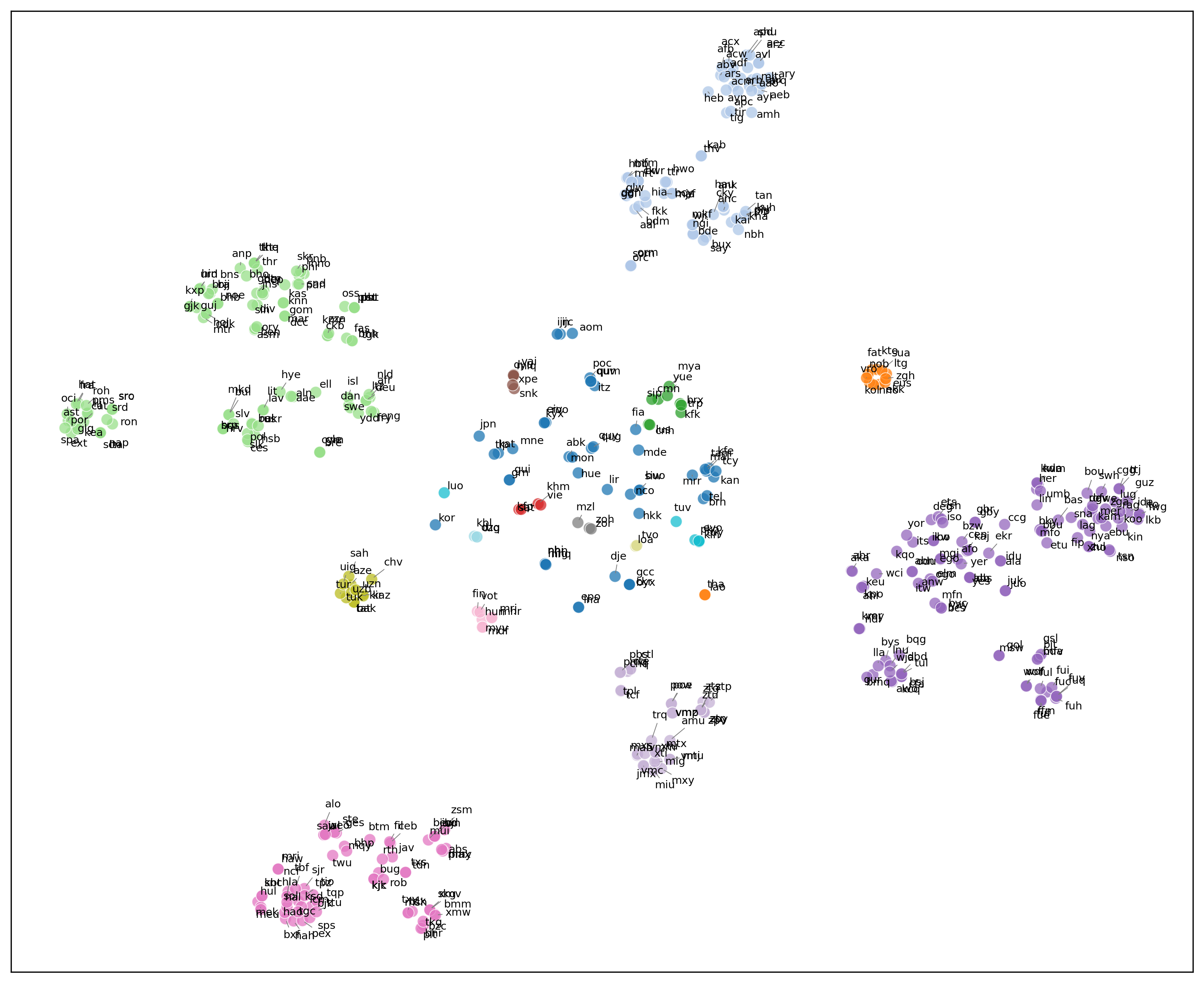}
        \caption{Cluster formation using the genetic vector.}
        \label{fig:app_genetic}
    \end{subfigure}
    
    \caption{Comparison of knowledge-guided clusters.}
    \label{fig:app_clusters_comparison_vector}
\end{figure*}
In this section, we provide detailed visualizations of the clustering results for each explicit grouping strategy, as illustrated in Figures~\ref{fig:app_clusters_comparison_embedding} and~\ref{fig:app_clusters_comparison_vector}. Languages assigned to the same group are indicated by the same color. Each plot is annotated with ISO 639-3 language codes following the Ethnologue\footnote{https://www.ethnologue.com/} convention.

\end{document}